\documentclass[]{fairmeta}

\usepackage{float}      
\usepackage{placeins}   
\usepackage{graphicx}

\usepackage{xcolor}
\definecolor{metablue}{RGB}{24, 119, 242}

\title{\textcolor{metablue}{\textbf{GLIMPSE}}: Real-Time Text Recognition and Contextual Understanding for VQA in Wearables}

\author[1]{Akhil Ramachandran}
\author[1]{Ankit Arun}
\author[1]{Ashish Shenoy}
\author[1]{Abhay Harpale}
\author[1]{Srihari Jayakumar}
\author[1]{Debojeet Chatterjee}
\author[1]{Mohsen Moslehpour}
\author[1]{Pierce Chuang}
\author[1]{Yichao Lu}
\author[1,\dagger]{Vikas Bhardwaj}
\author[1,\dagger]{Peyman Heidari}

\affiliation[1]{Meta Reality Labs}
\contribution[\dagger]{Joint last author}

\abstract{
Video Large Language Models (Video LLMs) have shown remarkable progress in understanding and reasoning about visual content, particularly in tasks involving text recognition and text-based visual question answering (Text VQA). However, deploying Text VQA on wearable devices faces a fundamental tension: text recognition requires high-resolution video, but streaming high-quality video drains battery and causes thermal throttling. Moreover, existing models struggle to maintain coherent temporal context when processing text across multiple frames in real-time streams. We observe that text recognition and visual reasoning have asymmetric resolution requirements - OCR needs fine detail while scene understanding tolerates coarse features. We exploit this asymmetry with a hybrid architecture that performs selective high-resolution OCR on-device while streaming low-resolution video for visual context. On a benchmark of text-based VQA samples across five task categories, our system achieves 72\% accuracy at 0.49x the power consumption of full-resolution streaming, enabling sustained VQA sessions on resource-constrained wearables without sacrificing text understanding quality.
}

\date{\today}
\correspondence{First Author at \email{akhilram@meta.com}}

\begin{document}

\maketitle

\section{Introduction}
\label{section:intro}


Wearable devices are rapidly proliferating, evolving from niche accessories into pervasive, everyday computing platforms ~\citep{kim-etal-2024-healthllm,ferrara2024largelanguagemodelswearable,sensorlm-2025}. In parallel, Large Language Models (LLMs) have become widely adopted and demonstrate strong capabilities in understanding, reasoning, and dialog across diverse domains. This convergence motivates systems that integrate sensor-rich Wearables with the reasoning capacity of LLMs to enable continuous, context-aware interaction in real-world settings ~\citep{lumos_2024}. 

A central application in this space is visual question answering (VQA) ~\citep{ge-etal-2024-lmmvqa, antol2015vqa}, where users pose questions about their surroundings and receive timely, contextually accurate answers ~\citep{lin-etal-2024-videollava,chen-etal-2025-vqaguider}. Multimodal LLMs (MMLLMs), particularly Video LLMs, are increasingly capable of processing and integrating visual and auditory signals with language understanding~\citep{radford2021clip, li2023blip2}, supporting hands-free assistance that perceives, maintains context, and reasons over what the user sees and hears ~\citep{llava-video-2024,ge-etal-2024-lmmvqa}.

\begin{figure*}[htbp]
    \centering
    \includegraphics[width=\textwidth]{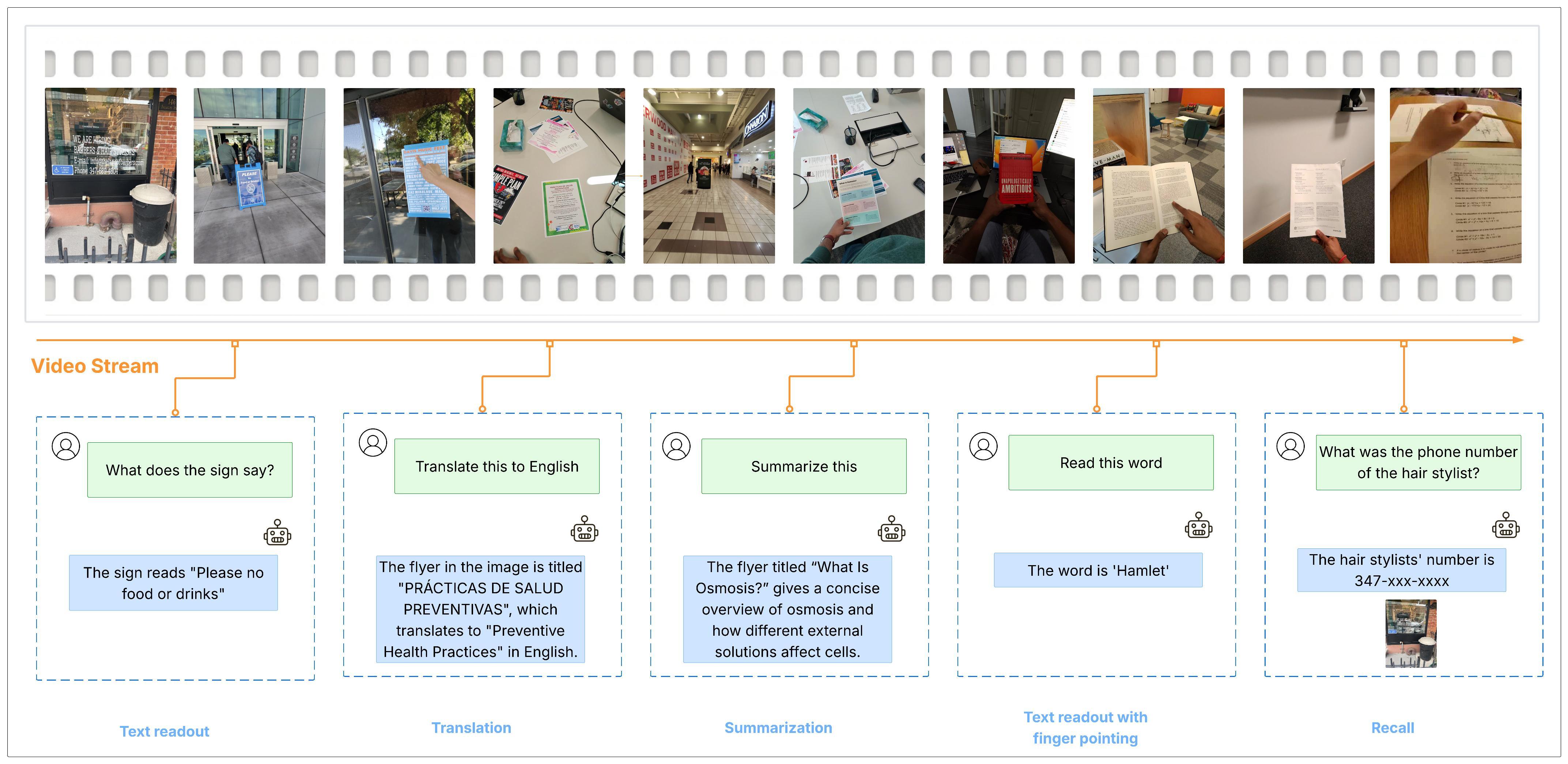}
    \caption{Illustrative wearable VQA scenarios: a user poses natural-language questions about text in their environment (e.g.,
   signs, labels, menus) and receives real-time contextual answers through the GLIMPSE pipeline.}
    \label{fig:ph1}
\end{figure*}

Textual input processing is especially critical for VQA because much of the actionable information in everyday scenes is encoded as text—names, quantities, prices, instructions, schedules, warnings, and identifiers. Unlike generic visual attributes, these tokens carry precise, compositional semantics that often determine the answer exactly (e.g., a gate number, dosage, or time). A robust VQA system must therefore treat text as a first-class signal. Specifically, it should:
\begin{itemize}
    \item Detect and normalize relevant spans,
    \item Track updates over time,
    \item Preserve uncertainty,
    \item Fuse textual evidence with the linguistic query and visual context,
\end{itemize}
so that answers remain well-supported, disambiguated, and faithful ~\citep{avinash2024realtime}. ~\cref{fig:ph1} illustrates a typical user interaction scenario where our proposed system processes real-time visual input to provide contextual text-based VQA.

The conventional approach ~\citep{chen2024videollmonlineonlinevideolarge,huang2024vincirealtimeembodiedsmart} offloads perception and reasoning to server-side infrastructure. The wearable captures high-fidelity video (and optionally audio) and streams it to a backend where compute-intensive pipelines execute. A scene-text subsystem performs detection and recognition, canonicalizing spans with spatial–temporal metadata and confidence scores. 

The effectiveness of this server-offload pipeline depends on access to high-quality frames. Scene-text recognition quality is directly related to overall video quality ~\citep{yin2016text}—including resolution, sharpness, and stream bitrate. Lower resolution and bitrates diminish fine-detail fidelity and text legibility, leading to character confusions and degraded span parsing, which ultimately reduces answer quality ~\citep{text-recognition-eval-2024,mlkit-mobile-ocr-2024}. Table \ref{tab:llmres}  shows that text recognition accuracy increases with video resolution and sufficient bitrate to preserve detail, underscoring the need to maintain stream quality throughout the pipeline. However, this tension motivates a key observation: text recognition and visual reasoning have asymmetric resolution requirements. While OCR accuracy degrades sharply below 12MP (from ~89\% to ~20\% at 3MP), scene understanding for VQA (object detection, layout comprehension, visual context) remains effective at much lower resolutions. This asymmetry suggests that text and video need not be transmitted at the same fidelity.

\begin{table}
  \centering
  \begin{tabular}{lc}
    \hline
    \textbf{Video resolution} & \textbf{Accuracy} \\
    \hline
    \verb|3MP (1536 x 2048 pixels)|     & {19.80\%}           \\
    \verb|5MP (1944 x 2592 pixels)|     & {56.44\%}           \\
    \verb|12MP (3024 x 4032 pixels)|    & {89.04\%}           \\\hline
  \end{tabular}
  \caption{Text recognition accuracy degrades sharply at lower resolutions. Word-level accuracy drops from 89\% at 12MP to under 20\% at 3MP,
  motivating the need for high-resolution input to the OCR stage}
  \label{tab:llmres}
\end{table}

Continuous streaming of high-quality video from a wearable device imposes substantial power and energy demands, dominated by (i) high-frame-rate capture and encoding of high-quality frames and (ii) intralink transfer of large-bandwidth video streams, as illustrated in Table \ref{tab:videopower}. The cumulative load shortens operating lifetime, elevates device temperature, and renders long-running sessions impractical. To remain within power budgets, systems typically reduce resolution or bitrate—choices that directly diminish the text fidelity required by downstream recognition and, consequently, VQA performance.

At first glance, running models locally avoids high-bandwidth video offload. In practice, wearable devices operate under tight compute, memory, and battery limits that make sustained on-device inference difficult ~\citep{acm-ondevice-2024,awesome-llms-on-device-2024}. Real-time video further imposes strict latency constraints—per-frame inference must complete within ~33 ms at 30 fps, or outputs become stale and temporally inconsistent (Table \ref{tab:latency}). Effective on-device operation therefore requires careful budgeting rather than a simple shift to fully local processing ~\citep{al2024latency, al2025intelligent}.

Recent research marks a shift toward hybrid architectures that distribute large language model (LLM) workloads between edge devices and server infrastructure, moving beyond the limitations ~\citep{ahmad2026visionlanguagemodelsedgerealtime} of traditional cloud-offload and fully on-device approaches. These systems employ a variety of strategies, like selective task partitioning ~\citep{edgecloudai} - executing detail-dependent operations locally on resource-constrained devices, while offloading context integration and advanced reasoning to remote servers or edge nodes;  edge device sharding ~\citep{zhang2024edgeshard} partitioning the LLM into shards and deploying on distributed devices etc. In this paper we further explore the task partitioning strategy to develop a power optimized text recognition system to support VQAs for Wearables.

To summarize, our contributions are as follows:

\begin{itemize}
    \item We identify that text recognition and visual reasoning have asymmetric resolution requirements, and propose a hybrid architecture that exploits this: selective high-resolution OCR on-device, low-resolution video streaming for visual context, and server-side fusion for VQA.
    \item We introduce a three-stage Smart Frame Selection pipeline (blur detection, ROI/text detection, similarity filtering) that reduces OCR workload by 67.7\% while preserving text fidelity.
    \item We design an OCR Session Manager that handles temporal alignment, deduplication, and retrieval of sparse OCR payloads for prompt construction.
    \item On our text-based VQA benchmark, our system achieves 72\% accuracy at 0.49x the power of full-resolution streaming.
\end{itemize}

\begin{table*}
  \centering
  \begin{tabular}{llll}
    \hline
    \textbf{Capture resolution} & \textbf{Capture Rate} & \textbf{Encoding Bitrate} & \textbf{Wearable Device Power} \\
    \hline
    \verb|12MP (3024 x 4032 pixels)|     & {30 fps}  & {3 Mbps} & {x}           \\
    \verb|3MP (1536 x 2048 pixels)|     & {30 fps}  & {1 Mbps} & {(0.83)x}           \\
    \verb|3MP (1536 x 2048 pixels)|     & {12 fps}  & {1 Mbps} & {(0.65)x}           \\
    \verb|3MP (1536 x 2048 pixels)|     & {2 fps}  & {500 Kbps} & {(0.49)x}           \\
    \hline
  \end{tabular}
  \caption{End-to-end energy consumption under different video streaming configurations. Power is reported relative to the 12MP/30fps baseline
  (1.0x). Reducing resolution from 12MP to 3MP and frame rate from 30fps to 2fps lowers power to 0.49x, establishing the target power budget for
  GLIMPSE's low-resolution streaming path.}
  \label{tab:videopower}
\end{table*}

\section{Detailed Architecture}

\begin{figure*}[htbp]
    \centering
    \includegraphics[width=\textwidth]{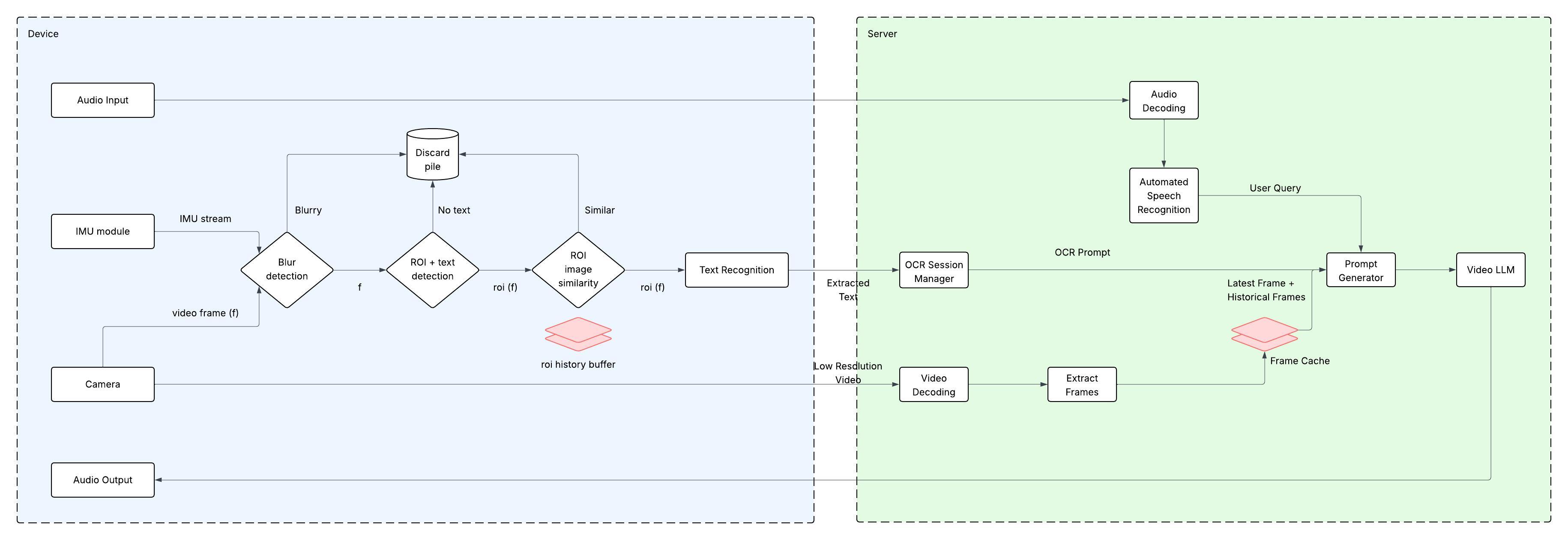}
    \caption{Overview of the GLIMPSE hybrid architecture. On-device components (Smart Frame Selection, high-resolution OCR) process video locally and
  transmit sparse text payloads, while a low-resolution video stream is sent to the server where the OCR Session Manager and Video LLM perform fusion
  and reasoning for VQA.}
    \label{fig:design}
\end{figure*}

\subsection{On-Device Components}

The on-device architecture centers around video frame processing at 2 frames per second (fps) captured from the Wearable camera to minimize power. Frames are processed at an original resolution of 12MP (3024 x 4032 pixels) in raw Y8 grayscale format. The core challenge addressed by this architecture is the need to perform real-time text recognition while maintaining acceptable power consumption and thermal characteristics throughout VQA sessions. Multiple specialized components work in concert to achieve intelligent frame selection and efficient processing.

\subsubsection{Smart Frame Selection}

Processing every single video frame with text recognition leads to significant power consumption and latency. For example, a naive implementation for VQA would require 3,600 OCR inferences (2~fps $\times$ 1,800~seconds for a 30-minute session), resulting in major power and thermal issues. To address this, the system employs a Smart Frame Selection pipeline that intelligently selects frames for processing, maintaining high-quality text understanding while reducing resource usage.

The Smart Frame Selection system uses a multi-stage pipeline with rejection at each stage, aiming to reduce OCR inferences to less than 1~fps (over 2$\times$ reduction). Empirical results show frame selection rates of 10--20\% in non-text-dense videos, achieving approximately 5$\times$ reduction in power consumption.

\paragraph{Blur Detection}

Model-based blur detection serves as the first stage in the Smart Frame Selection pipeline. Its principal function is to filter out video frames that are affected by blur, thereby ensuring that only frames with sufficient sharpness progress to subsequent stages of text processing. Blurriness can be introduced by multiple factors including motion, and light conditions. Figure \ref{fig:blur} demonstrates the impact of blur in video quality and text recognition. Early elimination of low-quality frames is crucial for both effectiveness and efficiency: blurred frames not only degrade the performance of text recognition algorithms but also waste computational resources if allowed to continue through the pipeline. The blur detection model uses the following features:

\begin{figure}[t]
  \centering
  \includegraphics[width=0.40\linewidth]{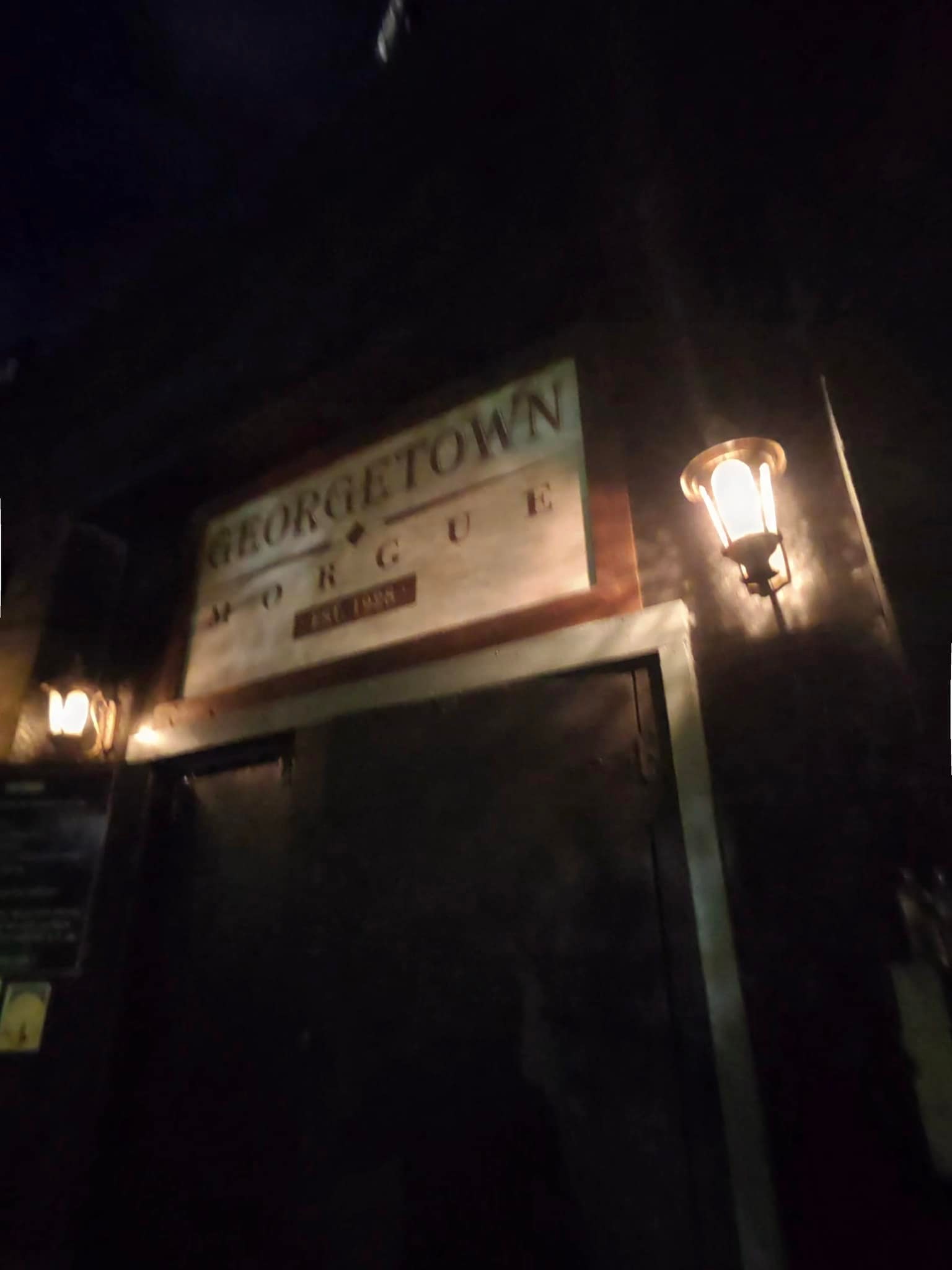}
  \hspace{0.03\linewidth}
  \includegraphics[width=0.40\linewidth]{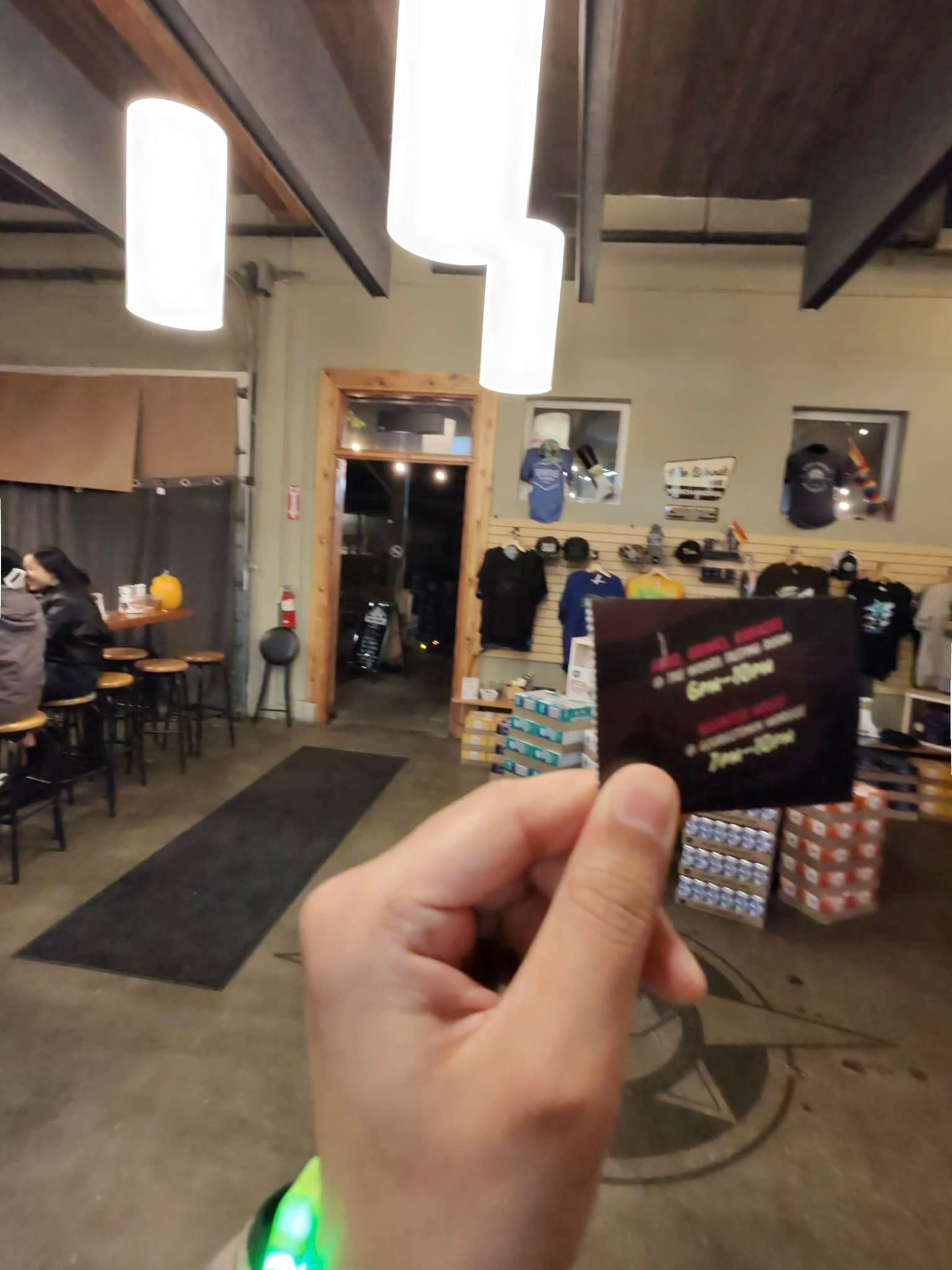}
  \caption{Impact of blur on text legibility in wearable video frames. (Left) Motion blur from rapid device movement renders text unreadable. (Right)
  Low-light conditions produce underexposed, noisy frames. The blur detection stage uses IMU-derived motion energy and camera exposure time to reject
  such frames before OCR processing.}
  \label{fig:blur}
\end{figure}

\begin{itemize}
    \item \textbf{IMU-derived motion energy:} The Euclidean norm of inertial measurement unit (IMU) sensor data (gyroscope and accelerometer), computed across the frame's exposure period to quantify device movement during capture ~\citep{ekdahl2023inertial}.
    \item \textbf{Camera exposure time:} The exposure duration reported by the camera, as longer exposures increase susceptibility to motion blur.
\end{itemize}

These features, reflecting both dynamic (motion) and static (exposure) capture conditions, are used by the model as input signals. The representation is processed by a lightweight decision-tree-based classifier, which has been trained on annotated datasets containing both sharp and blurred images. Based on these inputs, the classifier determines the blur status of each incoming video frame and enables prompt rejection of low-quality candidates.

The blur detection model was trained using samples taken with a Wearable device, consisting of IMU and exposure information, which were manually annotated as either blurry or not blurry. This targeted, small-scale dataset enabled the development of a decision-tree-based classifier tailored to the specific characteristics of wearable device capture scenarios.

\paragraph{Region of Interest and Text Detection}

The Region of Interest (ROI) and text detection model constitutes the second stage of the Smart Frame Selection pipeline. The key motivation is that, in real-world scenarios, the area containing relevant text often represents only a small portion of the overall image, even when the object is close to the camera. Processing the entire high-resolution frame for text is computationally intensive and can quickly exceed the wearable device’s resource constraints and latency bounds, while simply reducing the image size risks making the text too small to be accurately recognized.

\begin{figure*}[t]
  \includegraphics[width=0.22\linewidth]{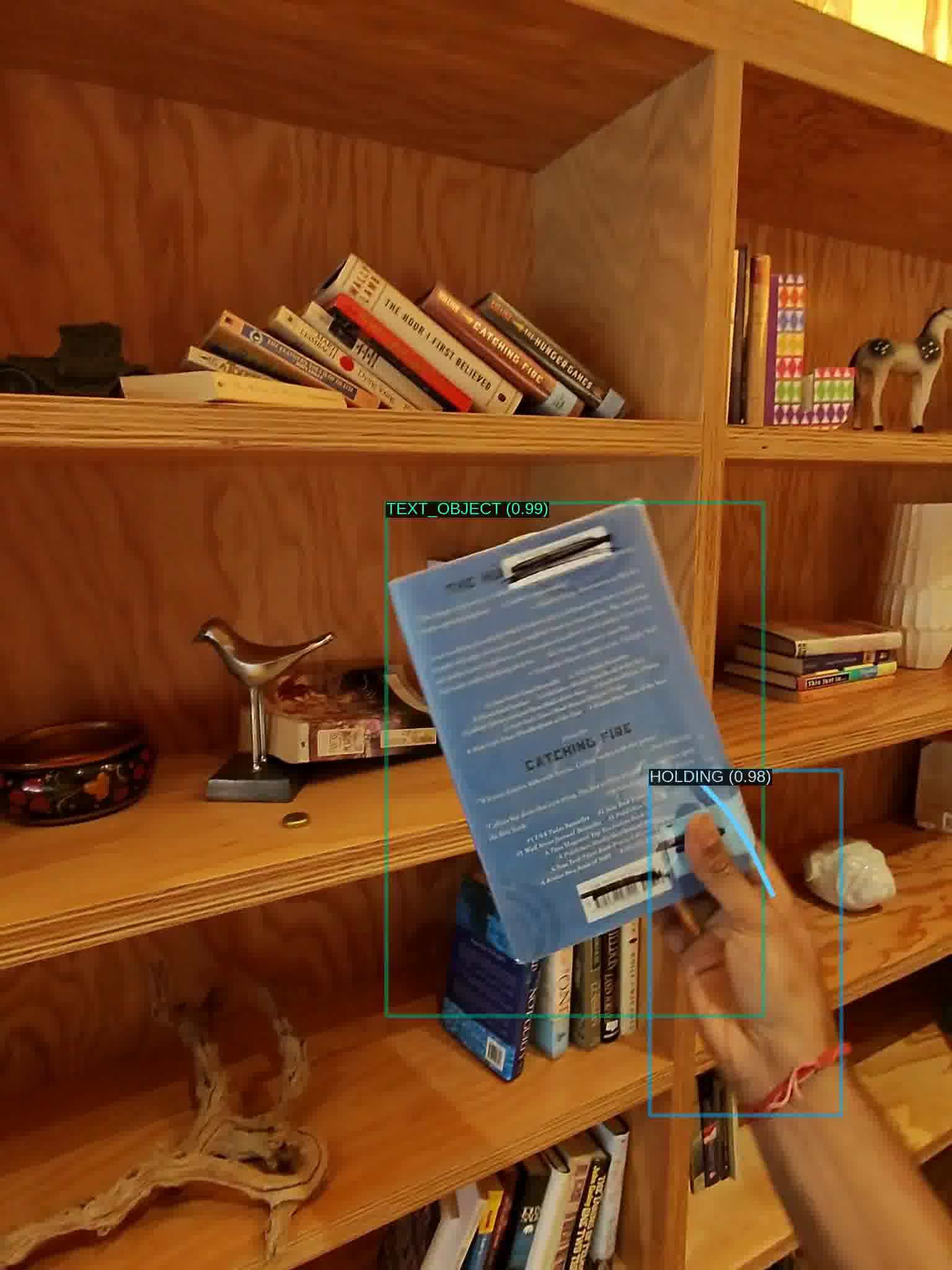} \hfill
  \includegraphics[width=0.22\linewidth]{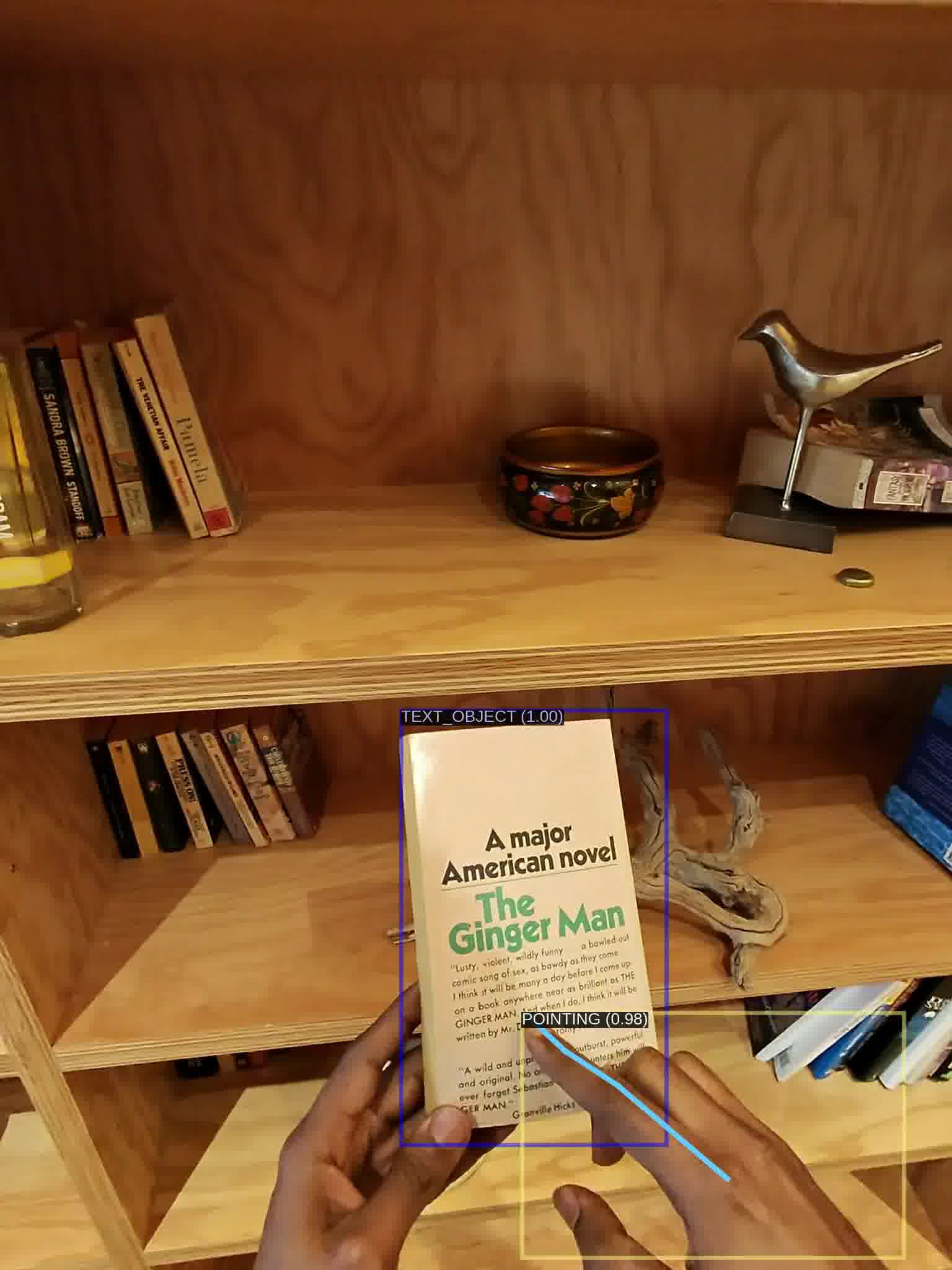} \hfill
  \includegraphics[width=0.22\linewidth]{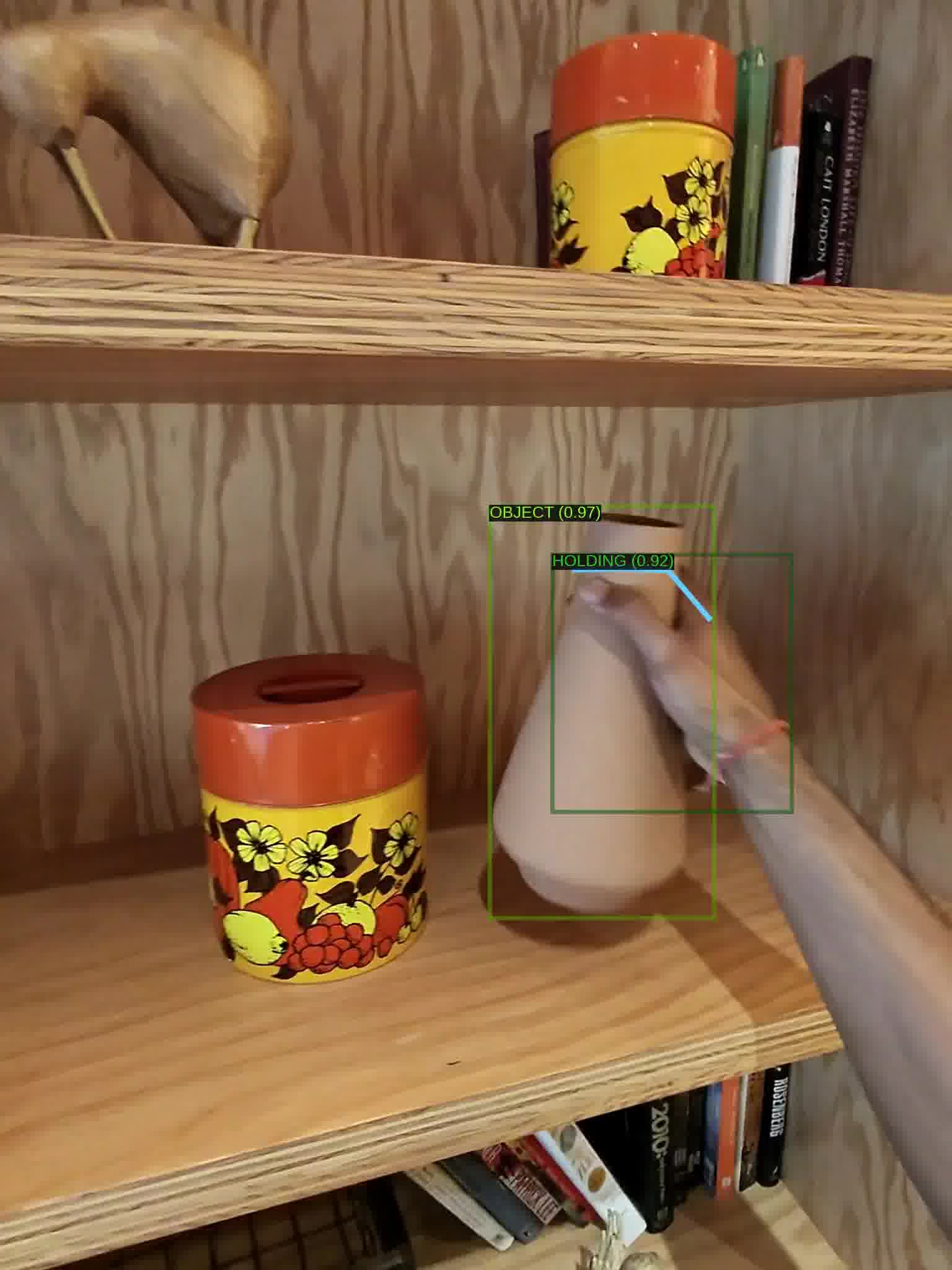} \hfill
  \includegraphics[width=0.22\linewidth]{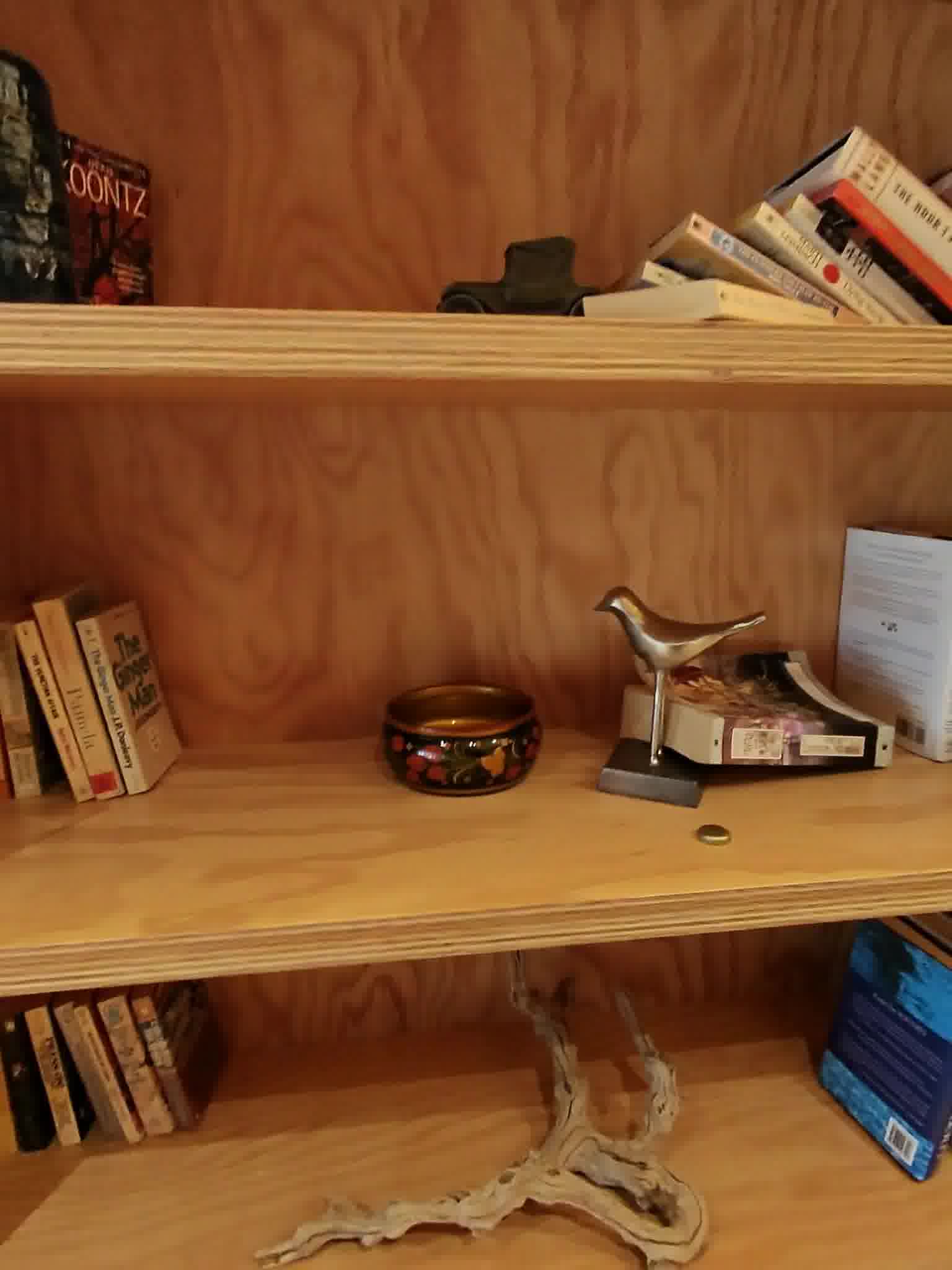} 
  \caption {ROI and text detection outputs across four interaction scenarios. (a) Hand-holding a text-bearing object: the model crops the detected
  text region at full resolution for OCR. (b) Finger-pointing at a text area: the pointed-at region is selected as the primary ROI. (c) Hand-holding a
  non-text object: the model correctly identifies the object but skips OCR since no text is present. (d) No salient region detected: the frame is passed
   without text processing. Bounding boxes and keypoints are shown as model outputs.}
  \label{fig:haptic}
\end{figure*}

Additionally, user interactions such as holding or pointing provide context for ROI selection, and filtering ROIs without text avoids unnecessary processing like text detection algorithms on full-resolution video frames, further optimizing both power consumption and processing speed.

The system utilizes a unified multi-task model for ROI and text detection, optimized for on-device efficiency and accuracy. This model operates directly on down-sampled grayscale image buffers at a 3MP resolution, a choice that balances computational efficiency with sufficient spatial detail for reliable detection. By consolidating the detection of objects, hand interactions, and text-containing regions into a single inference, the solution achieves a total processing latency of approximately 70 milliseconds per video frame. Figure \ref{fig:haptic} demonstrates how the model identifies the ROI and text on video frames.

For each video frame, the model performs the following steps:
\begin{itemize}
    \item Object detection and classification, identifying bounding boxes for categories such as hand-pointing, hand-holding, other hand interactions, and objects likely to contain text.
    \item For detected hands, estimation of keypoints for the index finger joints and tip, enabling precise inference of pointing gestures.
    \item Application of confidence thresholds to all detections; the object with the highest confidence score is selected as the primary region of interest for subsequent text recognition.
\end{itemize}

The model was trained using 80K in-the-wild text images annotated with salient regions, and 20K images with hand holding or finger pointing. To reduce false positives caused by accidental hands, we included 10K images with a hand that is neither holding nor pointing as hard negatives in our training data.

\subsection{Server-Side Components}

The server-side pipeline operates primarily on OCR payloads emitted on-device rather than raw video, ingesting time-aligned text records and applying specialized enrichment models to improve textual quality and downstream VQA performance. Concretely, server models perform autocorrection, object-aware grouping that constrains grouping and ordering within instance-segmented object boundaries and post-processing such as normalization, deduplication, temporal consolidation, entity linking, and confidence calibration, producing structured entries for session-scoped retrieval and prompt assembly. By decoupling enrichment from continuous high-resolution video streaming and avoiding redundant frame transfers, the system reduces uplink bandwidth and wearable power consumption while still leveraging high-capacity server models to refine the extracted text and strengthen grounding for the Video LLM’s answers

\subsubsection{OCR session manager}
The OCR Session Manager (OSM) is a key component in our hybrid wearable-server architecture, responsible for managing the lifecycle and retrieval of text recognition results from real-time video streams. The OCR outputs obtained from devices are sparse and irregularly distributed across the video timeline. OSM is designed to organize these outputs, maintain temporal alignment, and provide robust retrieval of text data for any requested timestamp, ensuring seamless integration with server-side processing and VQA.
Due to selective frame processing, the OCR payloads managed by OSM can be categorized as follows:
\begin{enumerate}
    \item \textbf{Receiving OCR Payloads:} As video frames are processed, SFS determines which frames are sent for OCR. For each frame, OSM receives one of five payload types:
    \begin{itemize}
        \item \textbf{Text detected and OCR executed:} Contains transcription results.
        \item \textbf{No text detected:} Empty payload.
        \item \textbf{Scene similar to previous:} OCR skipped, empty payload.
        \item \textbf{Blurry frame:} OCR skipped, empty payload.
        \item \textbf{Resource constraint:} OCR not run, empty payload.
    \end{itemize}
    \item \textbf{Temporal Sorting:} All payloads are maintained in a sorted order by their frame timestamps, regardless of arrival order.
    \item \textbf{Include at Least One Nonempty OCR:} For any batch of requested timestamps, OSM ensures that at least one nonempty OCR payload is included in the result, even if most frames are empty or skipped. This guarantees that meaningful text is available for downstream tasks.
    \item \textbf{Multiframe Aggregation and Exemplar Selection:} OSM groups similar OCR payloads (based on textual content or embeddings) across multiple frames. Within each group, a single exemplar is selected—typically the payload with the longest, highest-quality text. This reduces redundancy and ensures that the most informative OCR is used for prompt generation. Payloads with explicit user selection (e.g., finger-pointing) are treated as singletons and not aggregated, preserving user intent.
    \item \textbf{Deduplication by Timestamp:} For frames marked as similar, OSM deduplicates by associating the valid OCR result with the latest timestamp in the group. Only the freshest occurrence in each group is returned for batch requests.
    \item \textbf{Selection Handling:} For frames with explicit user selection, OSM ensures the corresponding OCR payload is fresh and not skipped, even if SFS would otherwise skip it.
    \item \textbf{Retrieval Logic:} When an OCR result is requested for a specific timestamp $T$:
    \begin{itemize}
        \item If a valid OCR payload (type 1 or 2) exists for $T$, it is returned as-is.
        \item If only empty/skipped payloads (types 3, 4, 5) exist for $T$, OSM returns the latest valid OCR payload before $T$.
        \item If no payload exists for $T$, OSM returns the latest valid OCR payload before $T$.
    \end{itemize}
    For fallback cases, the returned OCR is updated with the requested timestamp to ensure correct alignment.
    \item \textbf{Prompt Generation:} OSM prepares the OCR payloads for downstream AI by:
    \begin{itemize}
        \item Including only the most relevant, non-duplicate OCRs.
        \item Enriching prompts with scene context flags (e.g., blurry, upside-down, cropped, poor lighting).
        \item Suppressing stale selection information, retaining only the latest selection.
    \end{itemize}
\end{enumerate}

\subsection{Video LLM}

The multimodal large language model (MMLLM) powering VQA is built on a Mixture-of-Experts (MoE) architecture ~\citep{DBLP:journals/corr/ShazeerMMDLHD17}, which activates only a subset of parameters per token to improve computational efficiency while maintaining model capacity. This design reduces serving latency, making it well-suited for real-time wearable applications. Rather than relying on a specific model instantiation, the system is designed to be compatible with any sufficiently capable multimodal backbone that supports scalable training and efficient inference.

\subsection{Inference Context}

The inference system employs a multi-faceted approach to context management, commencing with a refined \textbf{Frame Selection Strategy} that dynamically captures visual context via three components: \textit{Pre-Query Frames} (uniform sampling prior to speech), \textit{In-Query Frames} (high-priority capture during speech, favoring high-resolution data), and \textit{Historical Frames} (context from previous conversation turns). During \textbf{Prompt Construction}, all selected context—including frames, conversation history, and OCR data—is organized into \textit{PromptComponent} objects, which are assembled and chronologically ordered using the heuristics. Extracted text is fed into the model via \textit{OCR in the prompt} using a standard format that includes timestamps and the text content, along with special features like QR code extraction and quality warnings. This text is further governed by specific prompts for readout and translation which are added as preambles to dynamically instruct the model, such as "Read this word by word, spell out license plates character by character" or "Translate this word by word into <language>". Critical optimizations are applied across the pipeline, including frame sampling techniques like prompt caching via \textbf{Warmup Requests} to prefill the KV cache ~\citep{shoeybi2020megatronlmtrainingmultibillionparameter} for reduced time-to-first-token latency, and \textbf{OCR Deduplication} using Multiframe Aggregation with FAISS similarity search to ensure efficiency.

\section{Experiments}

\subsection{Datasets}
For the evaluation dataset, we collected 112 videos with 1-3 questions per video. We then used OCR and video data with human annotators to write answers to the questions. This resulted in 208 question/answer pairs that cover different types of text understanding situations, as shown in Table \ref{tab:per-category}. The dataset includes translation tasks, mathematical reasoning from text, short-form text readout (signs, labels, menus), long-form text readout (articles, documents), and analysis and summarization of content (questions requiring text comprehension beyond simple readout).

\begin{table}[t]
\centering
\begin{tabular}{lcc}
\hline
\textbf{Category} & \textbf{Size} & \textbf{Accuracy} \\
\hline
\verb|Short readout| & 98 & 76\% \\
\verb|Translation| & 30 & 72\% \\
\verb|Analysis and Summary| & 53 & 70\% \\
\verb|Long readout| & 20 & 63\% \\
\verb|Math reasoning| & 7 & 57\% \\
\hline
\textbf{Overall} & \textbf{208} & \textbf{72\%} \\
\hline
\end{tabular}
\caption{VQA accuracy broken down by task category. Short readout tasks (signs, labels) achieve the highest
  accuracy (76\%), while math reasoning (57\%) is most challenging due to error propagation from symbol misrecognition. Overall system accuracy is 72\%.}
\label{tab:per-category}
\end{table}

\subsection{Results}

The videos along with OCR payloads extracted from them were passed to the Video LLM along with questions, and the responses were evaluated against ground truth. An LLM judge ~\citep{zheng2023judging} was used to compare ground truth with responses from the VQA system, classifying each into three categories:
\begin{enumerate}
    \item \textbf{Correct:} Response matches closely with the ground truth in terms of correctness and level of detail.
    \item \textbf{Poor:} Response is correct and related to the question but does not directly answer it.
    \item \textbf{Incorrect:} Response is incorrect in detail and/or unrelated to the question asked.
\end{enumerate}

\subsubsection{Comparison with Baselines}
To contextualize our results, we compare against two baseline configurations: (1) \textbf{Server-Full}, which streams 12MP video at 30fps and performs all OCR and reasoning server-side, and (2) \textbf{Server-Low}, which streams 3MP video at 2fps to match our system's power budget. As shown in Table~\ref{tab:baseline}, Server-Low achieves only 41\% accuracy due to degraded OCR quality at low resolution (see Table~\ref{tab:llmres}). Our hybrid approach recovers near-full accuracy (72\% vs.\ 74\%) while maintaining the power efficiency of Server-Low, demonstrating the effectiveness of decoupling high-resolution OCR from low-resolution video streaming.

\subsubsection{Per-Category Analysis}
Table~\ref{tab:per-category} presents accuracy broken down by task category. Short readout tasks achieve the highest accuracy (76\%), as these involve recognizing brief, clearly-visible text such as signs, labels, and menus. Translation tasks perform well (72\%), benefiting from high-quality OCR that preserves source text fidelity. Analysis and summarization tasks show moderate accuracy (70\%), as these require both accurate text extraction and higher-level comprehension.
Long readout tasks prove more challenging (63\%), as longer text sequences provide more opportunities for OCR errors to accumulate and affect downstream reasoning. Math reasoning achieves the lowest accuracy (57\%), reflecting the compounded difficulty of precise symbol recognition and multi-step logical inference—small OCR errors (e.g., misreading ``6'' as ``8'') propagate to incorrect final answers.

\subsubsection{Smart Frame Selection Ablation}
Table~\ref{tab:sfsefficiency} shows the contribution of each Smart Frame Selection stage. The blur filter provides a modest reduction (2\%), primarily filtering motion-blurred frames during head movement. The text content filter contributes the largest reduction (38.1\%), rejecting frames where no text-containing region of interest is detected. The similarity filter provides an additional 29.6\% reduction by skipping frames with unchanged text content. Cumulatively, the three-stage pipeline reduces OCR workload by 67.7\% while preserving text fidelity for downstream VQA.

\begin{table}
  \centering
  \begin{tabular}{lc}
    \hline
    \textbf{Category} & \textbf{Percentage} \\
    \hline
    \verb|Correct|     & {72.05\%}           \\
    \verb|Poor|     & {7.35\%}           \\
    \verb|Incorrect|    & {20.59\%}           \\\hline
  \end{tabular}
  \caption{Distribution of VQA response quality as judged by an LLM evaluator. Responses are classified as Correct (matching ground truth in detail),
   Poor (partially relevant but not directly answering the question), or Incorrect (factually wrong or unrelated).}
  \label{tab:results}
\end{table}

\begin{table}[t]
\centering
\begin{tabular}{lccc}
\hline
\textbf{System} & \textbf{VQA Acc.} & \textbf{Power} \\
\hline
\verb|Server-Full (12MP)| & 74\% & 1.00x \\
\verb|Server-Low (3MP)| & 41\% & 0.49x \\
\verb|Ours| & \textbf{72\%} & \textbf{0.49x} \\\hline
\end{tabular}
\caption{Comparison with baseline configurations. Server-Full streams 12MP video at 30fps; Server-Low streams 3MP at 2fps. Our hybrid approach matches Server-Full accuracy at Server-Low power consumption.}
\label{tab:baseline}
\end{table}

\section{Conclusion}
We presented a hybrid architecture for text-based VQA on wearable devices, motivated by a key observation: text recognition and visual reasoning have asymmetric resolution requirements - OCR needs high-resolution input while scene understanding tolerates lower fidelity. By performing selective high-resolution OCR on-device and streaming low-resolution video for visual context, our system achieves 72\% VQA accuracy at 0.49x the power consumption of full-resolution streaming. A three-stage smart frame selection pipeline reduces OCR workload by 67.7\%, enabling sustained sessions without thermal throttling. To maintain coherent temporal context despite sparse and irregularly-timed OCR outputs, we introduced an OCR Session Manager (OSM) that organizes payloads by timestamp, aggregates similar text across frames to select high-quality exemplars, and provides robust retrieval for any query time—ensuring the Video LLM always receives temporally-aligned text even when frames are skipped or filtered. Our work demonstrates that decoupling text and visual streams, rather than treating video as a monolithic signal—offers a practical path toward power-efficient multimodal understanding on resource-constrained devices.

\section {Limitations}
While our benchmark of 208 samples was designed to capture realistic wearable text-VQA scenarios across five task categories, broader evaluation on established benchmarks (e.g., TextVQA~\citep{singh2019textvqa}, DocVQA~\citep{mathew2021docvqa}) would provide additional validation. The current system prioritizes Latin-script languages, with multilinguality with a larger vocab and handwritten text support as natural extensions.

\clearpage
\newpage
\bibliographystyle{assets/plainnat}
\bibliography{paper}

\newpage

\beginappendix
\FloatBarrier

\begin{table}[H]
  \centering
  \begin{tabular}{lc}
    \hline
    \textbf{Word count}  & \textbf{Recognition latency} \\
    \hline
    \verb|0 words|    & {341 ms}     \\
    \verb|30 words|        & {396 ms}      \\
    \verb|100 words|      & {1188 ms}       \\
    \verb|1000 words|     & {4976 ms}       \\\hline
  \end{tabular}
  \caption{On-device OCR inference latency as a function of the number of words detected in a frame, establishing practical limits on per-frame processing budgets and motivating the Smart Frame Selection pipeline.}
  \label{tab:latency}
\end{table}

\begin{table}[H]
  \centering
  \begin{tabular}{lll}
    \hline
    \textbf{Stage} & \textbf{Video frame count} & \textbf{Percentage change} \\
    \hline
    \verb|Camera stream|     & {37.40 M}   & {-}        \\
    \verb|After Blur Filter|     & {36.63 M}   & {-2.0\%}        \\
    \verb|After Text Content Filter|     & {23.13 M}   & {-38.1\%}        \\
    \verb|After Similarity Filter|     & {12.09 M}   & {-67.7\%}        \\\hline
  \end{tabular}
  \caption{Cumulative frame reduction through the three-stage Smart Frame Selection pipeline, measured on production video streams (37.4M frames).}
  \label{tab:sfsefficiency}
\end{table}

\begin{table}[H]
  \centering
  \begin{tabular}{llc}
    \hline
    \textbf{Scenario} & \textbf{Word count} & \textbf{Wearable device power} \\
    \hline
    \verb|12 fps, no text recognition| & \verb| -| & x \\
    \verb|2 fps, no text recognition| & \verb| -| & (0.85)x \\
    \verb|12 fps, text recognition on all frames| & 
      \begin{tabular}{l}
        \verb|0 words| \\
        \verb|30 words| \\
        \verb|100 words|
      \end{tabular} &
      \begin{tabular}{l}
        (1.42)x \\
        (1.68)x \\
        (1.88)x
      \end{tabular} \\
    \verb|12 fps, text recognition on sampled frames at 2fps| &
      \begin{tabular}{l}
        \verb|0 words| \\
        \verb|30 words| \\
        \verb|100 words|
      \end{tabular} &
      \begin{tabular}{l}
        (1.31)x \\
        (1.54)x \\
        (1.77)x
      \end{tabular} \\
    \verb|2 fps, text recognition on all frames| &
      \begin{tabular}{l}
        \verb|0 words| \\
        \verb|30 words| \\
        \verb|100 words|
      \end{tabular} &
      \begin{tabular}{l}
        (0.95)x \\
        (1.06)x \\
        (1.08)x
      \end{tabular} \\
    \verb|2 fps, smart frame selection on 12MP input frame| &
      \begin{tabular}{l}
        \verb|0 words| \\
        \verb|30 words| \\
        \verb|100 words|
      \end{tabular} &
      \begin{tabular}{l}
        (1.05)x \\
        (1.11)x \\
        (1.19)x
      \end{tabular} \\
    \verb|2 fps, smart frame selection on down-scaled 3MP frame| &
      \begin{tabular}{l}
        \verb|0 words| \\
        \verb|30 words| \\
        \verb|100 words|
      \end{tabular} &
      \begin{tabular}{l}
        (0.91)x \\
        (0.94)x \\
        (0.96)x
      \end{tabular} \\
    \hline
  \end{tabular}
  \caption{Detailed power consumption breakdown across video capture and OCR configurations on the wearable device, reported relative to a 12fps
  no-OCR baseline (1.0x). Key findings: (1) reducing frame rate from 12fps to 2fps saves ~15\% power even without OCR; (2) OCR on all frames at 12fps
  adds 42-88\% overhead depending on word count; (3) Smart Frame Selection on down-scaled 3MP frames keeps total power under 0.96x for scenes with up to
  100 words, validating the chosen operating point.}
  \label{tab:power}
\end{table}

\end{document}